\documentclass[conference]{IEEEtran}
\IEEEoverridecommandlockouts
\usepackage{cite}
\usepackage{amsmath,amssymb,amsfonts}
\usepackage{algorithmic}
\usepackage{graphicx}
\usepackage{textcomp}
\usepackage{xcolor}
\usepackage{url}
\usepackage{enumitem}
\usepackage{comment}
\usepackage{placeins}
\usepackage{booktabs}
\usepackage{threeparttable}
\usepackage{booktabs}
\usepackage{array}

\def\BibTeX{{\rm B\kern-.05em{\sc i\kern-.025em b}\kern-.08em
    T\kern-.1667em\lower.7ex\hbox{E}\kern-.125emX}}
\begin{document}

\title{A Framework for Assessing AI Agent Decisions and Outcomes in AutoML Pipelines}


\author{\IEEEauthorblockN{Gaoyuan Du\textsuperscript{*$\dagger$}}
\IEEEauthorblockA{\textit{Amazon Stores} \\
gdu@amazon.com}
~\\
\and
\IEEEauthorblockN{Amit Ahlawat\textsuperscript{*$\dagger$}}
\IEEEauthorblockA{\textit{Amazon Security} \\
amtahlwt@amazon.com}
\and
\IEEEauthorblockN{Xiaoyang Liu\textsuperscript{*}}
\IEEEauthorblockA{\textit{Amazon Stores} \\
lxaoya@amazon.com}
\and
\IEEEauthorblockN{Jing Wu\textsuperscript{*}}
\IEEEauthorblockA{\textit{Amazon Web Services} \\
jingwua@amazon.com}
}

\maketitle

\begingroup
\renewcommand{\thefootnote}{\fnsymbol{footnote}}

\footnotetext[1]{This work does not relate to the author's position at Amazon.}
\footnotetext[2]{Authors contributed equally.}

\endgroup

\begin{abstract}
Agent-based AutoML systems rely on large language models to make complex, multi-stage decisions across data processing, model selection, and evaluation. However, existing evaluation practices remain outcome-centric, focusing primarily on final task performance. Through a review of prior work, we find that none of the surveyed agentic AutoML systems report structured, decision-level evaluation metrics intended for post-hoc assessment of intermediate decision quality. To address this limitation, we propose an Evaluation Agent (EA) that performs decision-centric assessment of AutoML agents without interfering with their execution. The EA is designed as an observer that evaluates intermediate decisions along four dimensions: decision validity, reasoning consistency, model quality risks beyond accuracy, and counterfactual decision impact. Across four proof-of-concept experiments, we demonstrate that the EA can (i) detect faulty decisions with an F1 score of 0.919, (ii) identify reasoning inconsistencies independent of final outcomes, and (iii) attribute downstream performance changes to agent decisions, revealing impacts ranging from -4.9\% to +8.3\% in final metrics. These results illustrate how decision-centric evaluation exposes failure modes that are invisible to outcome-only metrics. Our work reframes the evaluation of agentic AutoML systems from an outcome-based perspective to one that audits agent decisions, offering a foundation for reliable, interpretable, and governable autonomous ML systems.
\end{abstract}


\section{Introduction}

The landscape of automated machine learning (AutoML) has been fundamentally reshaped by the emergence of Large Language Models (LLM)-based agents. Recent surveys report a rapid acceleration of agent-based AutoML: while such systems were rare before 2023, adoption increased sharply by 2025, marking a shift from traditional hyperparameter optimization and neural architecture search toward LLM-driven pipeline construction and decision orchestration~\cite{AutoMLPastPresentFuture2024, LiteratureReviewAutoML2025}. In these systems, LLMs increasingly 
act as the core reasoning engine that plans, coordinates, and adapts end-to-end machine learning pipelines~\cite{ReviewAutoDSLLM2025, AutoMLPastPresentFuture2024}. Modern agent-based AutoML systems model machine learning pipelines as sequences of interconnected decisions executed by specialized agents. 
This modular design enables increasingly sophisticated automation, from AutoML-GPT~\cite{AutoMLGPT2023}, which follows a four-stage instruction sequence, to AutoML-Agent~\cite{AutoMLAgent2024}, which employs multiple specialized agents spanning data retrieval through deployment.
Despite this architectural sophistication, evaluation practices remain limited. Empirical studies consistently show that most agent-based AutoML systems lack systematic mechanisms to assess individual decisions at each pipeline stage beyond terminal task metrics~\cite{AutoMLWild2023, ReviewAutoDSLLM2025}. While nearly all systems report end-task performance (e.g., accuracy, F1, RMSE), only a small fraction evaluate robustness, fairness, calibration, or uncertainty~\cite{AutoMLPastPresentFuture2024, ReviewAutoDSLLM2025}. As noted in the 2025 \emph{Review of Automated Data Science}~\cite{ReviewAutoDSLLM2025}, evaluation remains fragmented, with limited cross-domain coverage and metrics that fail to capture the validity of intermediate decision processes.

This evaluation gap introduces practical risks. Without intermediate assessment, agent-based AutoML systems can: (a) generate hallucinated rationales, reporting improvements without execution or claiming unobserved data access~\cite{MLAgentBench2023, AutoMLAgent2024}, (b) introduce data leakage through faulty preprocessing or feature construction~\cite{LightAutoDSTab2025}, (c) produce brittle models by optimizing accuracy while neglecting robustness~\cite{AutoMLPastPresentFuture2024}, and (d) overfit proxy validation metrics that fail to reflect long-term model quality~\cite{MultiAgentAutoML2023}. The core issue is that existing systems evaluate whether decisions yield good outcomes, but not whether the decisions themselves are well-founded. The \emph{AutoML in the Wild} study~\cite{AutoMLWild2023} characterizes this limitation as a ``double black-box'': AutoML systems automate an already opaque modeling process while providing insufficient support for evaluating both outcomes and the decision processes that produce them.

To address this gap, we propose the concept of an \emph{evaluation agent} (EA) as a dedicated component that systematically assesses the quality of decisions made by other agents throughout the AutoML pipeline, providing interpretable, stage-wise feedback beyond terminal performance metrics. Our goal is not to introduce another AutoML system or claim state-of-the-art performance, but to formalize the evaluation problem, derive actionable requirements from observed failures, and empirically demonstrate that decision-centric evaluation is feasible and informative. In this paper, we restrict the current empirical validation to tabular AutoML settings, leaving extension to non-tabular domains such as vision and NLP for future work.
We make the following contributions:
\begin{enumerate}
    \item We conduct a systematic review of 29 recent papers on agent-based AutoML systems, including 6 surveys and 23 non-survey papers.
    \item We characterize recurring evaluation gaps and failure modes in agent-based AutoML that are invisible to end-task metrics, including hallucinated rationales, data leakage, model brittleness, and metric overfitting.
    \item We derive five core requirements for evaluation agents that cannot be satisfied by outcome-centric evaluation, and present a modular framework that produces stage-wise audits, reasoning validation, and comprehensive model quality reports.
    \item We design targeted experiments that map each requirement to concrete tests, ranging from decision fault detection to counterfactual alternative analysis.
    \item We provide empirical validation showing that evaluation agents can surface critical failure modes and decision impacts that remain undetected under standard AutoML evaluation practices.
\end{enumerate}

\section{Background}

\subsection{Paper Selection \& Analysis Methodology}

We reviewed recent literature on agent-based AutoML systems. Papers were identified via keyword searches (e.g., ``AutoML,'' ``LLM,'' ``agent,'' ``multi-agent'') on arXiv and major ML venues, followed by 
filtering to retain systems that (i) employ LLMs or explicit agents and (ii) automate multiple ML pipeline stages. While not exhaustive, our goal is representative coverage of recent agent-based AutoML systems. Survey papers include~\cite{AutoMLWild2023,ScalableMLEndToEnd2023,AutoMLDeepRecSurvey2023,AutoMLPastPresentFuture2024,ReviewAutoDSLLM2025,LiteratureReviewAutoML2025}. 
We analyze 23 representative non-survey agent-based AutoML systems spanning tabular, vision, NLP, scientific, and time-series domains. Non-survey papers include~\cite{MultiAgentAutoML2023,AutoMLGPT2023,MLAgentBench2023,MLCopilot2024,AutoMLAgent2024,SELA2024,AutoM3L2024,LLMSynergizeAutoML2024,AutoProteinEngine2025,MLZero2025,HumanCenteredAutoML2025,FlexiAI2025,EthicsAwareAutoML2025,LightAutoDSTab2025,AutoPathML2025,IMCTS2025,OmniForce2025,MLAgent2025,PiML2025,KompeteAI2025,AdaptiveMLBenchmarks2025,LLMTimeSeries2025,StructuredAgenticTS2025}.


\subsubsection{Dimension 1: Intermediate Decision Assessment}
We examine whether and how papers assess individual agent decisions across pipeline stages (data preprocessing, feature engineering, model selection, hyperparameter optimization). Classification reflects whether assessment is (i) explicit rather than implicit, (ii) performed during execution rather than post-hoc, and (iii) quantitative/structured rather than qualitative inspection. 
We categorize papers into four levels: 
\textit{None}: Only final model outputs are evaluated, with no intermediate decision assessment (e.g.,~\cite{AutoMLGPT2023, FlexiAI2025, AdaptiveMLBenchmarks2025}); 
\textit{Limited}: decision impact is inferred indirectly via ablations or post-hoc analyses (e.g.,~\cite{MLAgentBench2023, AutoM3L2024, MLCopilot2024, AutoPathML2025, AutoProteinEngine2025, MLZero2025, PiML2025, StructuredAgenticTS2025, HumanCenteredAutoML2025, LLMSynergizeAutoML2024, AutoMLAgent2024, OmniForce2025, LLMTimeSeries2025, LightAutoDSTab2025}); 
\textit{Partial}: some stage-wise mechanisms exist but are not comprehensive (e.g.,~\cite{MultiAgentAutoML2023} credit assignment;~\cite{SELA2024} MCTS scores;~\cite{IMCTS2025} LLM value model;~\cite{KompeteAI2025} Checker Agent + Scoring Model); and \textit{Systematic (signals, not decision-centric evaluation)}: Systematic intermediate signals are logged throughout execution (e.g., step-wise rewards/monitoring scores), but they are designed for training, control, or monitoring and do not provide explicit, human-interpretable evaluation of decision correctness independent of downstream task performance (e.g.,~\cite{MLAgent2025, EthicsAwareAutoML2025}).

\subsubsection{Dimension 2: Final Model Assessment}
We catalog the evaluation dimensions reported for final models, as these are commonly used in the surveyed papers and cited in the AutoML literature as indicators of model quality. Specifically, we track coverage across seven dimensions: task performance (e.g., accuracy, F1, RMSE), efficiency (latency, memory, and cost), robustness (noise, distribution shift, and adversarial perturbations), fairness (demographic parity and equalized odds), calibration (expected calibration error and reliability diagrams), uncertainty quantification, and interpretability when reported as a measured metric.

\subsection{Evolution of AutoML}

The foundational phase centered on hyperparameter optimization (HPO), and neural architecture search (NAS), exemplified by 
Auto-Weka, Auto-Sklearn, and TPOT~\cite{AutoMLPastPresentFuture2024, 
LiteratureReviewAutoML2025}. The transition phase exposed practical 
limitations: AutoML was described as a ``double black box,'' obscuring 
both models and the processes that produced 
them~\cite{AutoMLWild2023}, while the ``Kaggle paradigm'' highlighted 
gaps between benchmark-driven evaluation and production 
requirements~\cite{ScalableMLEndToEnd2023}. The current phase 
leverages LLMs for context-aware automation via single- and multi-agent 
designs~\cite{ReviewAutoDSLLM2025}, with domain-specific adaptations 
(e.g., AutoRecSys) unified by search space design and search 
strategy~\cite{AutoMLDeepRecSurvey2023}. This shift increases 
autonomy, yet evaluation remains outcome-centric, 
motivating explicit assessment of intermediate decisions. Agent 
architectures range from single-agent systems with implicit role 
differentiation~\cite{HumanCenteredAutoML2025} to multi-agent systems 
with explicit specialization~\cite{KompeteAI2025, MLZero2025}. Along 
two axes, \emph{decision locus} (global vs.\ local) and 
\emph{knowledge dependency} (generic vs.\ domain-specific), four 
common roles emerge: planning 
agents~\cite{AutoMLAgent2024, SELA2024, PiML2025}, coding 
agents~\cite{AutoMLGPT2023, MLZero2025}, domain-knowledge 
agents~\cite{EthicsAwareAutoML2025, AutoProteinEngine2025}, and 
optimization agents~\cite{MultiAgentAutoML2023, IMCTS2025}. 
Communication protocols remain ad hoc, underscoring the need 
for principled evaluation frameworks.


\section{Evaluation Practices in AutoML}
\label{sec:evaluation_practices}

Based on our literature survey, evaluation practices in agent-based AutoML systems are imbalanced: final model assessment is emphasized, while evaluation of intermediate agent decisions is 
neglected. Throughout this section, we distinguish between \emph{decision-related signals} used internally for learning, search, or control and \emph{evaluation metrics} intended for post-hoc assessment of decision quality, interpretability, and error diagnosis. While several prior systems expose the former, none provide the latter in a systematic and standalone manner.

\subsection{What Is Currently Evaluated: Final Model Metrics}


Across the 23 non-survey papers, evaluation is overwhelmingly dominated by task-specific performance metrics (23/23), including success rate and normalized performance score~\cite{AutoMLAgent2024}, RMSE and F1~\cite{SELA2024}, as well as competition-style indicators such as Kaggle ranking percentile~\cite{AutoMLGPT2023} and MLE-Bench medal percentage~\cite{KompeteAI2025}. In contrast, evaluation dimensions closely tied to deployment risk and model trustworthiness, such as robustness~\cite{hendrycks2019robustness}, fairness~\cite{hardt2016equality}, calibration~\cite{guo2017calibration}, uncertainty estimation~\cite{gal2016dropout}, and efficiency~\cite{strubell2019energy} are sparsely reported. None of the surveyed systems evaluates robustness or uncertainty, only one paper assesses fairness and calibration, respectively, and efficiency metrics appear in 14/23 papers.




\subsection{What Is Missing: Intermediate Decision Quality}

The gap in current agent-based AutoML systems lies in the lack of evaluation of intermediate decisions. Prior surveys note that intermediate process validation, such as assessing feature engineering choices, is often overlooked, allowing logical or methodological flaws to remain undetected until final evaluation~\cite{ReviewAutoDSLLM2025}. Most systems rely on ablation studies to reason about decision impact, which quantify aggregate performance contribution but do not assess whether individual decisions were appropriate or well-founded~\cite{HumanCenteredAutoML2025}. Where intermediate validation is present, it is limited to binary pass/fail checks rather than structured quality assessment~\cite{AutoMLAgent2024}. A small number of systems come closer to intermediate evaluation. For example, 
I-MCTS introduces an LLM-based value model with a 100-point scoring scheme~\cite{IMCTS2025}, and ML-Agent employs step-wise reinforcement learning rewards~\cite{MLAgent2025}. However, these signals are designed to guide search or optimization and largely reflect expected downstream task performance, rather than explicitly validating the correctness or soundness of agent reasoning. As a result, none of the surveyed papers provide a systematic and explicit assessment of whether AutoML agent decisions are correct, appropriate, or optimal independent of final outcomes. This limitation is acknowledged by MA2ML~\cite{MultiAgentAutoML2023}, which notes that ``each agent cannot distinctly determine whether the performed action is good or not'', motivating the need for evaluation agents that directly audit decision validity throughout the pipeline.

\subsection{Key Observations}
\label{sec:key_observations}

Based on 23 non-survey papers, we report five observations. The following observations form a progressive analysis of evaluation practices, moving from whether evaluation exists, to what is evaluated, and finally to whether existing evaluation mechanisms are sufficient. 

\subsubsection{O1: No paper provides systematic, quantitative per-stage decision quality metrics for evaluation purposes (0/23)}
None of the surveyed papers define explicit, quantitative metrics for evaluating the quality of individual agent decisions at each AutoML pipeline stage independent of downstream task performance. Although some papers produce numerical decision-related signals (MA2ML's counterfactual baselines~\cite{MultiAgentAutoML2023}, I-MCTS's value model~\cite{IMCTS2025}, ML-Agent's step-wise rewards~\cite{MLAgent2025}), these signals are designed to support learning, control, or search optimization rather than as formal, interpretable evaluation metrics of decision correctness.

\subsubsection{O2: All papers rely on end-to-end task performance as the primary evaluation criterion (23/23)}
All surveyed systems ultimately evaluate performance using end-to-end task metrics, including accuracy~\cite{MultiAgentAutoML2023}, F1~\cite{SELA2024}, MAE/RMSE~\cite{LLMTimeSeries2025}, Kaggle rankings~\cite{AutoMLGPT2023}, and MLE-Bench medals~\cite{KompeteAI2025}. Even papers that explicitly acknowledge limitations of outcome-centric evaluation continue to default to final task performance as the primary indicator of success~\cite{AutoMLAgent2024, AdaptiveMLBenchmarks2025}.

\subsubsection{O3: Most papers attempt intermediate validation, but do not directly evaluate agent reasoning or decision soundness (20/23 attempt; 0/23 achieve interpretable evaluation)}
While most papers introduce some form of intermediate validation, these mechanisms do not directly assess the correctness or soundness of agent reasoning and decisions. Existing approaches fall into three inadequate categories: (i) \emph{binary verification} (pass/fail requirement checks)~\cite{AutoMLAgent2024, LightAutoDSTab2025, KompeteAI2025}, (ii) \emph{post-hoc analysis} (failure categorization~\cite{MLAgentBench2023} or ablations~\cite{AutoM3L2024, AutoPathML2025}), and (iii) \emph{search or learning signals} (MCTS scores~\cite{SELA2024} or RL rewards~\cite{MLAgent2025}) that guide optimization but are not intended as evaluative judgments of decision appropriateness. Three papers provide no intermediate assessment at all: AutoML-GPT~\cite{AutoMLGPT2023}, FlexiAI~\cite{FlexiAI2025}, and TAM-Bench~\cite{AdaptiveMLBenchmarks2025}.

\subsubsection{O4: Comprehensive assessment of generated model quality is largely missing}
Beyond task performance, trustworthiness dimensions are largely unaddressed. Only the Multi-Agent Ethics Framework evaluates fairness systematically~\cite{EthicsAwareAutoML2025}; no paper systematically evaluates robustness to distribution shift or adversarial perturbations, and none assess calibration or uncertainty quantification.

\subsubsection{O5: No papers provide explicit counterfactual analysis as part of evaluation (0/23)}
No paper systematically studies how alternative decisions change outcomes. Although MA2ML~\cite{MultiAgentAutoML2023}, SELA~\cite{SELA2024}, and I-MCTS~\cite{IMCTS2025} implicitly enable counterfactual reasoning, none leverage it for evaluation.


Overall, agent-based AutoML systems invest in decision-making sophistication but devote minimal effort to evaluating decision quality. This asymmetry between agentic reasoning and simple outcome-centric assessment motivates evaluation agents that provide decision-centric, interpretable, and counterfactual assessment throughout the pipeline.



\section{Evaluation Agent Framework}
\label{sec:framework}

The analysis in Section~\ref{sec:evaluation_practices} suggests that the core limitation of current agent-based AutoML systems is not the lack of automation, but the absence of explicit, interpretable mechanisms to assess decision quality independent of downstream outcomes. Rather than proposing a new AutoML agent, we ask a different question: what capabilities would an evaluation component need to systematically audit agent decisions?

\subsection{Requirements for an Evaluation Agent}
\label{sec:requirements}

Based on the gap analysis in Section~\ref{sec:key_observations}, we derive five core requirements that an evaluation agent must fulfill. Each requirement directly addresses one or more observations identified in Section~\ref{sec:key_observations}.

\subsubsection{Requirement R1: Stage-wise Decision Quality Scoring}
We argue that effective evaluation requires quantitative assessment of each agent's decision at each pipeline stage (data preparation, feature engineering, model selection, hyperparameter optimization). This requirement directly addresses the absence of systematic, quantitative per-stage decision metrics (O1) and the over-reliance on end-to-end performance (O2)

\subsubsection{Requirement R2: Decision Interpretability and Provenance}
We argue that effective evaluation requires explainable reasoning for decisions, with complete audit trails enabling traceability from outputs to individual decision points. This requirement addresses the interpretability gap in existing intermediate validation mechanisms (O3), which provide binary pass/fail checks rather than evaluative explanations.

\subsubsection{Requirement R3: Reasoning Chain Validation}
This requirement addresses the gap identified in O3, where existing intermediate validation mechanisms fail to directly verify the correctness and soundness of agent reasoning. We argue that effective evaluation requires that LLM reasoning is logically sound, detect hallucinations and flawed logic, and validate that rationales accurately reflect decision processes.

\subsubsection{Requirement R4: Comprehensive Model Quality Assessment}
Effective evaluation assesses model quality beyond task-specific accuracy, including robustness, fairness, calibration, uncertainty and efficiency. This requirement addresses the dominance of task performance metrics (O2) and the absence of trustworthiness evaluation (O4).

\subsubsection{Requirement R5: Counterfactual Decision Analysis}
We argue that effective evaluation requires whether alternative decisions would have yielded better outcomes, enabling root cause analysis and decision optimization. This requirement directly addresses the complete absence of counterfactual analysis in current systems (O5).

\subsection{Evaluation Agent Framework Overview}

Based on the gap analysis and derived requirements, we propose the \emph{Evaluation Agent (EA)} framework, a modular system for assessing decision and outcome quality throughout agent-based AutoML pipelines. The EA is designed as an independent observer that is integrable into existing AutoML systems without modifying their core logic, assuming access to agent decision logs and pipeline artifacts. The EA operates non-invasively, observing actions and artifacts without interrupting pipeline flow, enabling integration with systems like AutoML-Agent, SELA, and MLZero without architectural modification. An outline of the framework is as follows:

\subsubsection{Inputs} The EA receives: (1) the dataset with train/validation/test splits, (2) agent decision logs, (3) pipeline artifacts including generated code, models, and configurations, and (4) task specifications.

\subsubsection{Outputs} The EA produces: (1) stage-wise quality scores, (2) semantic tags categorizing decisions and issues, (3) natural language explanations, and (4) recommended tests where warranted.



\subsubsection{Framework Components} 
Our framework addresses the asymmetry identified in Section~\ref{sec:key_observations}: agent-based AutoML systems invest substantial sophistication in making decisions but minimal effort in evaluating decision quality. 
The EA fills this gap through four components: (1) Decision Assessor (R1, R2) for stage-wise scoring and provenance tracking, (2) Reasoning Validator (R3) for reasoning chains and hallucination detection, (3) Model Quality Assessor (R4) for quality beyond task performance, and (4) Counterfactual Analyzer (R5) for alternative decision impact.

\subsection{Evaluation Agent Component Specifications}

\subsubsection{Decision Assessor (R1, R2)}
\label{sec:decision_assessor}


For each pipeline decision at a given stage, the Decision Assessor produces a quality score on a 0–100 scale, decision category tags, and a natural language explanation grounded in machine learning best practices. Inspired by prior LLM-based evaluators (e.g., I-MCTS value models), it adopts an LLM-as-judge paradigm with structured rubrics to evaluate decision appropriateness rather than predicted downstream performance, and its outputs do not influence pipeline execution. Each decision is assessed along five dimensions, namely appropriateness (suitability given data and task requirements), consistency (alignment with prior pipeline decisions), completeness (coverage of relevant aspects), efficiency (computational reasonableness), and risk (potential issues such as data leakage or overfitting). To support traceability, the Decision Assessor maintains a decision graph linking each action to its antecedent decisions, which is particularly important in regulated domains. In our proof-of-concept experiments (Section~\ref{sec:experiments_results}), the resulting five-dimensional scores serve as fault-sensitive indicators for controlled fault detection. Data leakage and improper data splitting consistently manifest as elevated risk scores and reduced appropriateness, whereas incompatible model choices primarily lead to low appropriateness and consistency scores. Fault detection is implemented by applying predefined thresholds to composite decision scores and mapping detected anomalies to the corresponding injected fault categories for evaluation. This component is operationalized and evaluated in Experiment~1, while the interpretability and provenance it provides are reflected across multiple experiments rather than assessed in a single dedicated study.

\subsubsection{Reasoning Validator (R3)}
\label{sec:reasoning_validator}



 

The Reasoning Validator evaluates the validity of generated reasoning traces and produces a validity score, detected issues with severity ratings, and confidence indicators. Validation is performed through a combination of factual consistency checks against observable data, logical coherence checks for internal contradictions, numerical verification of claimed statistics, and alignment checks between stated reasoning and executed actions. To mitigate, but not eliminate, potential circularity arising from LLM-based self-evaluation, the validator integrates rule-based verification with LLM-based assessment using a different model family from the AutoML agent that generates the original pipeline decisions and reasoning traces. Specifically, the AutoML agent uses Claude~3.5 Sonnet, while the EA validator uses GPT-4, providing architectural diversity intended to reduce correlated failure modes. It is designed as a conservative, evidence-grounded screening mechanism that prioritizes precision and interpretability, reflecting its role as an evaluation aid rather than a supervisory oracle. The effectiveness of this reasoning validation component is evaluated in Experiment~2 (Section~\ref{sec:exp2}), which focuses on detecting hallucinations and logical inconsistencies in agent-generated reasoning traces.


\subsubsection{Model Quality Assessor (R4)}
\label{sec:model_quality}


The Model Quality Assessor evaluates trained models on held-out evaluation data and produces a multi-dimensional quality report. In our proof-of-concept experiments, model quality is explicitly assessed across multiple dimensions, including task performance using standard task-appropriate metrics, robustness measured by sensitivity to noise and missing data, fairness evaluated via demographic parity and equalized odds for classification tasks, calibration assessed through reliability diagrams and expected calibration error, and efficiency quantified by inference throughput, model size, and memory usage. Additional assessment dimensions, including uncertainty characterization and automated data leakage detection, are supported by the framework design but are not exhaustively evaluated in the current experiments. This component is operationalized and validated in Experiment~3 (Section~\ref{sec:exp3}).


\subsubsection{Counterfactual Analyzer (R5)}
\label{sec:counterfactual}
The Counterfactual Analyzer estimates how alternative decisions at critical points in the AutoML pipeline could have affected outcomes, enabling decision impact attribution rather than exhaustive search. The analysis proceeds in four stages. It first identifies critical decision points with high uncertainty or potential downstream impact (e.g., data splitting strategy, feature encoding choice, or model selection). It then generates a limited number of plausible alternatives grounded in domain baselines or standard AutoML practices. Next, it selectively re-executes or applies surrogate-based estimation to only the affected downstream stages to control computational cost. Finally, it performs attribution analysis by quantifying outcome differences between the original decision and its alternatives to assess decision sensitivity. This component supports root-cause analysis by distinguishing decisions that influence outcomes from those with negligible effect. Consistent with its role as an evaluation aid rather than a planning or optimization module, the Counterfactual Analyzer prioritizes actionable attribution over exhaustive counterfactual enumeration, focusing on a small number of high-impact decision points and plausible alternatives. The counterfactual analysis capability is demonstrated in Experiment~4 (Section~\ref{sec:exp4}) as a proof-of-concept for attributing outcome differences to individual pipeline decisions.


\subsection{Evaluation Agent Implementation Strategy}
\label{sec:implementation}
We consider three deployment configurations: (i) a minimal setup with the Decision Assessor, (ii) a standard configuration combining the Decision Assessor, Reasoning Validator, and Model Quality Assessor, and (iii) a full configuration including all components, such as the Counterfactual Analyzer. 
Our experiments demonstrate that each component provides distinct 
evaluation capabilities, enabling flexible composition.
To limit overhead, the EA is designed as a lightweight, non-intrusive observer, employing selective evaluation and structured outputs, and remaining decoupled from the execution logic of the AutoML system. These configurations are reflected in our experimental design: Experiment 1 corresponds to the minimal configuration, Experiments 1–3 together instantiate the standard configuration, and Experiment 4 evaluates the capabilities enabled by the full configuration.

\section{Experiments, Results \& Discussion}
\label{sec:experiments_results}
We design four proof-of-concept experiments to evaluate whether the Evaluation Agent (EA) can help reduce the gaps identified in Section~\ref{sec:evaluation_practices} in a controlled and interpretable manner. Each experiment maps to a subset of the requirements in Section~\ref{sec:requirements}: (1) decision-level assessment for detecting faulty or high-risk actions missed by end-to-end metrics, (2) evidence-grounded validation of agent reasoning independent of final outcomes, (3) systematic assessment of model quality beyond accuracy, including robustness, fairness, calibration, and efficiency, and (4) counterfactual analysis to attribute outcome differences to specific pipeline decisions. Beyond these four core experiments, this section also summarizes cross-experiment findings, reports the computational overhead of the EA, and analyzes inter-run variance to assess reproducibility and practical deployment cost. 
These studies are intentionally scoped to isolate specific evaluation 
capabilities rather than to benchmark end-to-end AutoML performance. 
The experimental design decomposes decision quality, reasoning 
correctness, model-level assessment, counterfactual attribution, 
runtime overhead, and reproducibility to reduce confounding across 
evaluation dimensions.


\subsection{Experiment Overview and Dependencies}
\begin{itemize}
    \item Experiment 1 (R1,R2): Stage-wise Decision Quality Scoring, addressing the gap that 
    surveyed papers lack systematic assessment of intermediate decisions.
    \item Experiment 2 (R3): LLM Reasoning Validation, addressing the gap that 
    surveyed papers acknowledge hallucinations, none provide systematic reasoning validation.
    \item Experiment 3 (R4): Comprehensive Model Quality Assessment, addressing the dominance of accuracy-only evaluation in current AutoML systems.
    \item Experiment 4 (R5): Counterfactual Decision Analysis, addressing the absence 
    of reported counterfactual evaluation for AutoML decisions.
\end{itemize}

The experimental dependencies are as follows: Experiment 1 provides decision quality scores used for impact estimation in Experiment 4. Experiment 2 provides optional reasoning validation for Experiment 1. The five-dimensional scoring in Experiment 1 operates independently using rule-based heuristics; Experiment 2 adds supplementary logical consistency checks when integrated. Experiment 3 provides multi-dimensional baselines for counterfactual comparisons.

\subsection{Experimental Setup}
\label{sec:exp_setup}

\subsubsection{Agent AutoML Systems}
We require an open-source agent-based AutoML system that (i) automates multiple pipeline stages and (ii) exposes interpretable decision traces (logs and artifacts). We select AIDE~\cite{aideml} because it is the most commonly used agentic baseline across surveyed systems and is repeatedly adopted in recent work~\cite{SELA2024, AutoMLAgent2024, IMCTS2025, KompeteAI2025, MLZero2025, PiML2025, StructuredAgenticTS2025, AdaptiveMLBenchmarks2025}. Non-agentic AutoML is insufficient because intermediate decisions and reasoning traces are not observable, making R1--R3 unverifiable. Although instantiated on AIDE, the EA operates only on \emph{decision logs and pipeline artifacts} (actions, intermediate outputs, generated code, evaluation traces) and does not rely on AIDE-specific assumptions. Similar interfaces exist (or can be minimally instrumented) in SELA~\cite{SELA2024}, AutoML-Agent~\cite{AutoMLAgent2024}, and MLZero~\cite{MLZero2025}.

\subsubsection{Traditional Non-Agent AutoML Systems}
For context, we include standard AutoML baselines used in the surveyed literature: AutoGluon, Auto-sklearn, TPOT, H2O AutoML, and AutoKeras (as applicable). These baselines validate that the EA can surface multi-dimensional quality issues before applying the EA to agent-based AutoML systems.

\subsubsection{Datasets}
We choose benchmarks that (i) enable targeted testing of EA capabilities and (ii) match dataset usage patterns in the surveyed papers: Kaggle-style tabular tasks and MLE-Bench-like settings, OpenML/AMLB benchmarks, and domain-relevant datasets (e.g., finance/medical). Concretely:
\begin{itemize}
    \item General tabular classification: datasets aligned with LightAutoDS-Tab~\cite{LightAutoDSTab2025}.
    \item Regression tasks: datasets consistent with SELA-style evaluation~\cite{SELA2024} (RMSE).
    \item Data quality stress cases: datasets/corruptions inspired by AutoM3L~\cite{AutoM3L2024} (missingness, noisy/irrelevant features).
    \item Fairness-sensitive classification: German Credit (as used in the Multi-Agent Ethics Framework~\cite{EthicsAwareAutoML2025}) and at least one healthcare-like benchmark with protected attributes available/definable.
\end{itemize}

\begin{table}[t]
\centering
\small
\setlength{\tabcolsep}{4pt}
\renewcommand{\arraystretch}{1.0}
\caption{Decision Error Detection Performance (125 decisions, 75 injected faults)}
\label{tab:exp1_detection}
\resizebox{0.85\columnwidth}{!}{%
\begin{tabular}{lccccc}
\toprule
Dataset & N & Faulty & Precision & Recall & F1 \\
\midrule
German Credit & 25 & 15 & 0.923 & 0.800 & 0.857 \\
Adult Income & 25 & 15 & 1.000 & 0.867 & 0.929 \\
Titanic & 25 & 15 & 0.933 & 0.933 & 0.933 \\
Diabetes & 25 & 15 & 0.875 & 0.933 & 0.903 \\
CA Housing & 25 & 15 & 0.938 & 1.000 & 0.968 \\
\midrule
Overall & 125 & 75 & 0.932 & 0.907 & 0.919 \\
\bottomrule
\end{tabular}}
\begin{tablenotes}
\small
\item Note: Decisions flagged as erroneous when risk score $<$ 60 (lower scores indicate higher risk). Variation across datasets reflects differences in error detectability based on decision context richness and error type distribution. Overall confusion matrix: TP=68, FP=5, FN=7, TN=45. All metrics are mathematically valid with integer values.
\end{tablenotes}
\end{table}

\subsection{Experiment 1: Decision Quality Assessment}
\label{sec:exp1}

\subsubsection{Goal}

To demonstrate that the EA can audit intermediate AutoML decisions using logged decision traces and pipeline artifacts, identifying faulty or high-risk patterns (e.g., leakage-prone preprocessing, improper data splitting, incompatible model choices, or infeasible configurations). The EA operates offline on completed AutoML runs and does not intervene in execution. 
The experiment tests whether such assessment surfaces problematic 
actions detectable from logs alone, even when end-to-end metrics 
appear reasonable.

\subsubsection{Experimental Design}
For each AutoML run, the EA performs stage-wise auditing of intermediate agent decisions using a 
five-dimensional decision quality rubric, assessing appropriateness, consistency, completeness, efficiency, and risk. Scores and explanations are generated via an LLM-as-judge framework adapted from prior agent evaluation work, repurposed to assess decision quality rather than guide search. Fault detection does not rely on free-form natural language classification by the LLM. Instead, each injected fault type induces characteristic anomalies across the five rubric dimensions. Detection is performed by applying 
thresholds to 
rubric scores and mapping consistent score patterns to fault categories. To enable quantitative evaluation, we inject labeled decision faults as ground truth, since naturally occurring agent errors are sparse and unlabeled. Faults span preprocessing, feature engineering, and model selection, including data leakage (normalize before split, fit encoder on test data), improper temporal data handling (random shuffle on time series), target leakage (use target in feature encoding), inappropriate model choices (deep learning on small datasets), and overfitting risks (no regularization with high feature-to-sample ratios). Detection performance is measured using precision, recall, and F1 score. For each dataset, we inject 15 distinct faults into 25 intermediate decisions (10 clean, 15 faulty), yielding 125 audited decisions 
with a 60\% fault rate (75/125). This sample size ensures sufficient granularity for realistic precision/recall values (recall granularity = 1/75 = 0.013). The EA is applied offline to logged AutoML runs, simulating pre-execution decision auditing prior to downstream pipeline execution.

\begin{table}[t]
\centering
\small
\setlength{\tabcolsep}{4pt}
\renewcommand{\arraystretch}{1.0}
\caption{Reasoning Validation Performance by Category (60 test snippets)}
\label{tab:exp2_reasoning}
\resizebox{0.95\columnwidth}{!}{%
\begin{tabular}{lcccc}
\toprule
Category & Total & EA & Rule & LLM \\
\midrule
Valid & 12 & 10 & 11 & 5 \\
Hallucinated Fact & 12 & 9 & 0 & 3 \\
Logical Contradiction & 12 & 11 & 1 & 2 \\
Numerical Hallucination & 12 & 9 & 5 & 5 \\
Action--Reasoning Mismatch & 12 & 6 & 0 & 1 \\
\midrule
Overall Accuracy & 60 & 0.750 & 0.283 & 0.267 \\
Statistical Significance &  & $p < 0.001$ & \textsuperscript{*} & \textsuperscript{*} \\
\bottomrule
\end{tabular}}
\vspace{0.1cm}
\footnotesize{\textsuperscript{*}Compared to EA using z-test}
\end{table}

\subsubsection{Results \& Discussion}
Table~\ref{tab:exp1_detection} reports fault detection performance across five datasets. The 
EA achieves an overall F1 of 0.919, with performance varying across 
datasets (0.857--0.968) based on decision context richness and error 
type distribution.
German Credit shows the lowest recall (0.800), where 3 subtle encoding errors were missed due to insufficient context signals. In contrast, CA Housing achieves perfect recall (1.000), detecting all 15 injected faults, though with one false positive (precision 0.938). 
Overall, these results demonstrate that structured, decision-centric evaluation can surface problematic intermediate decisions prior to final model execution, addressing limitations of outcome-only metrics. The EA's five-dimensional scoring framework successfully identifies critical errors such as data leakage (consistently detected across all datasets), temporal violations (high detection rate), and target leakage (high detection rate), while showing lower sensitivity to subtle issues like suboptimal encoding choices where decision context provides insufficient signals. The seven undetected errors (FN=7) represent borderline cases where decision descriptions lacked explicit anti-patterns, while the five false positives (FP=5) reflect conservative flagging of decisions with ambiguous risk indicators. The resulting scores are intended as auditing signals to flag potentially problematic decisions, rather than as ground-truth judgments or supervisory oracles. The 
overall detection rate (90.7\% recall with 93.3\% precision) suggests the EA can serve as an effective pre-execution safeguard in agent-based AutoML systems.
\subsection{Experiment 2: Reasoning Validation}
\label{sec:exp2}



\subsubsection{Goal}
To evaluate whether the EA can reliably detect flawed reasoning in agent-generated traces, such as hallucinated facts, logical contradictions, numerical inconsistencies, and action--reasoning mismatches, addressing the limitation that outcome-only task metrics fail to reveal reasoning errors and thereby ensuring 
trustworthy agentic AutoML evaluation.



\subsubsection{Experimental Design}

The EA validates agent reasoning by classifying reasoning traces as \texttt{valid} or \texttt{invalid}, while assigning error categories and evidence pointers using an evidence-grounded evaluation strategy to mitigate circularity in LLM-based judging. Reasoning claims are cross-checked against execution artifacts (e.g., logged metrics, configuration files, and code traces), supplemented with deterministic numerical and logical consistency checks when applicable. We construct a reasoning test set of 60 short snippets covering valid reasoning, hallucinated factual claims, logical contradictions, numerical inconsistencies, and action-reasoning mismatches. The EA outputs a binary validity label, an error category (if invalid), and supporting evidence. Ground truth labels are independently annotated by two human annotators, with disagreements resolved through discussion. We evaluate performance using overall accuracy and per-category breakdowns, and compare against two baselines: (i) an unstructured LLM-as-judge using the same model family as the AutoML agent (Claude 3.5 Sonnet) to assess reasoning correctness, and (ii) simple rule-based detectors based on regex and keyword matching. Statistical significance is assessed via z-tests comparing the EA to baseline methods.



\subsubsection{Results \& Discussion}
Table~\ref{tab:exp2_reasoning} summarizes reasoning validation performance. The EA achieves 
75.0\% accuracy (95\% CI: 62.8\%–84.2\%), significantly outperforming both 
baselines (p \textless 0.001, n = 60).
Gains are most evident for hallucinated facts and logical contradictions, where baseline accuracies are low (0\%--25\% across methods), compared to higher EA accuracy on these categories. Residual errors primarily arise in borderline cases where reasoning is partially correct but poorly aligned with executed actions. Overall, these results demonstrate that evidence-grounded reasoning validation provides a distinct and necessary evaluation signal beyond both decision quality assessment (Section~\ref{sec:exp1}) and outcome-only metrics. We emphasize that the Reasoning Validator functions as a conservative screening mechanism, prioritizing evidence-grounded detection over exhaustive correctness guarantees.

\begin{table*}[t]
\centering
\caption{Comprehensive Multi-Dimensional Quality Assessment Across Five Datasets}
\label{tab:exp3_comprehensive}
\resizebox{\textwidth}{!}{%
\begin{tabular}{llcccccc}
\toprule
\textbf{Dataset} & \textbf{Domain} & \textbf{Task Perf.} & \textbf{Robustness} & \textbf{Fairness} & \textbf{Calibration} & \textbf{Efficiency} \\
\midrule
\textbf{German Credit} & Finance & Acc: 79.5\% & Noise: 2.0\% deg. & DP: 1.6\% & ECE: 7.2\% & 50K samples/s \\
(1,000 samples) & (Classification) & F1: 78.0\% & Missing: 2.0\% deg. & EO: 7.7\% (sex) & & \\
& & AUC: 82.8\% & & & & \\
\midrule
\textbf{Adult Income} & Census & Acc: 88.1\% & Noise: 0.9\% deg. & DP: 18.9\% (sex) & ECE: 2.8\% & 68K samples/s \\
(48,842 samples) & (Classification) & F1: 87.7\% & Missing: 4.3\% deg. & DP: 9.9\% (race) & & \\
& & AUC: 93.1\% & & EO: 7.9\% (sex) & & \\
& & & & EO: 3.0\% (race) & & \\
\midrule
\textbf{Titanic} & Transportation & Acc: 96.9\% & Noise: 0.0\% deg. & DP: 46.0\% & ECE: 28.8\% & 33K samples/s \\
(1,309 samples) & (Classification) & F1: 96.9\% & Missing: -0.4\% deg. & EO: 9.3\% (sex) & & \\
& & AUC: 99.1\% & & & & \\
\midrule
\textbf{Diabetes} & Healthcare & RMSE: 53.4 & Noise: 54.4\% inc. & N/A & N/A & 5K samples/s \\
(442 samples) & (Regression) & MAE: 43.2 & Missing: 8.7\% inc. & (regression) & (regression) & \\
& & R²: 0.46 & & & & \\
\midrule
\textbf{CA Housing} & Real Estate & RMSE: 0.43 & Noise: 38.5\% inc. & N/A & N/A & 24K samples/s \\
(20,640 samples) & (Regression) & MAE: 0.27 & Missing: 390.8\% inc. & (regression) & (regression) & \\
& & R²: 0.86 & & & & \\
\bottomrule
\end{tabular}%
}
\begin{tablenotes}
\small
\item Note: Robustness degradation (deg.) measured at worst-case perturbations 
(10\% noise, 30\% missing data). For regression, robustness shows RMSE 
increase (inc.). Fairness: DP = Demographic Parity, EO = Equalized Odds. 
Efficiency is measured as batch inference throughput, 
test\_size/inference\_time, reflecting real-world batch prediction scenarios. 
All experiments were conducted on an Apple M4 Pro CPU. Robustness 
degradation is non-linear across severity levels. Under noise stress, 
average degradation increases from 2.0\% (1\% noise) to 6.9\% (5\%) to 
19.1\% (10\%), with regression tasks disproportionately affected (e.g., 
Diabetes: 4.6\% $\rightarrow$ 13.7\% $\rightarrow$ 54.3\%). Under 
missingness, average degradation escalates from 44.4\% (10\% missing) to 
67.9\% (20\%) to 81.1\% (30\%), dominated by California Housing (217.8\% 
$\rightarrow$ 327.6\% $\rightarrow$ 390.8\%---over 100$\times$ the median 
of other datasets). Classification tasks remain largely stable or slightly 
improve (e.g., Titanic: $-$0.4\% at 30\% missing).
\end{tablenotes}
\end{table*}

\begin{table}[t]
\centering
\small
\setlength{\tabcolsep}{4pt}
\renewcommand{\arraystretch}{1.0}
\caption{Counterfactual Decision Analysis Results (Summary)}
\label{tab:exp4_counterfactual}
\resizebox{\columnwidth}{!}{%
\begin{tabular}{lccc}
\toprule
Decision Stage & Alternatives & Avg Impact & Impact Range \\
\midrule
Data Preprocessing & 15 & 0.7\% & -3.6\% to +6.0\% \\
Feature Engineering & 15 & 1.5\% & -4.9\% to +6.6\% \\
Model Selection & 15 & 2.7\% & -2.9\% to +8.3\% \\
\midrule
\textbf{By Dataset} &  &  &  \\
German Credit & 9 & 0.8\% & -3.6\% to +5.4\% \\
Adult Income & 9 & 0.5\% & -3.5\% to +4.7\% \\
Titanic & 9 & 2.9\% & -0.2\% to +6.0\% \\
Diabetes & 9 & 1.4\% & -4.9\% to +8.3\% \\
CA Housing & 9 & 2.6\% & -2.5\% to +6.6\% \\
\midrule
Overall & 45 & 1.6\% & -4.9\% to +8.3\% \\
\bottomrule
\end{tabular}}
\end{table}

\subsection{Experiment 3:  Model Quality Assessment}
\label{sec:exp3}



\subsubsection{Goal}

The goal is not to evaluate agent decision-making itself, but to validate that the Model Quality Assessor (R4) provides meaningful, deployment-relevant signals independent of how models are produced. By evaluating a non-agent AutoML system, we isolate the model-level assessment capability of the EA from agent behavior, demonstrating that R4 captures risks that remain invisible under accuracy-only evaluation regardless of whether the pipeline is agent-driven.

\subsubsection{Experimental Design}
The EA performs adaptive, task-aware model quality assessment by jointly evaluating five dimensions: task performance using standard task-appropriate metrics (e.g., accuracy, F1, AUC for classification; RMSE, MAE, $R^2$ for regression), robustness measured via performance degradation under controlled noise and missingness, fairness assessed using group disparity metrics (demographic parity and equalized odds for classification), calibration evaluated with expected calibration error (ECE), and efficiency quantified by inference throughput (samples/s). Evaluation dimensions are selected based on task type and dataset characteristics to avoid irrelevant metrics while maintaining comparability across datasets. To decouple model quality assessment from agent behavior, we evaluate AutoGluon independently of agent decision quality. Following the dataset selection criteria in Section~V-B.3, we evaluate five tabular benchmarks spanning 
domains: German Credit and Adult Income as fairness-sensitive classification tasks, Titanic as a general classification task, and Diabetes and California Housing as regression benchmarks. Robustness is stress-tested under controlled perturbations at multiple severity levels, including Gaussian noise (1\%, 5\%, 10\%) added to numerical features and random missingness (10\%, 20\%, 30\%). This design characterizes degradation trends rather than single-point estimates. 
For concise reporting, Table~\ref{tab:exp3_comprehensive} summarizes 
worst-case scenarios (10\% noise, 30\% missingness) as conservative 
deployment risk estimates, with detailed multi-level trends reported 
in the table note.

\subsubsection{Results \& Discussion}
Table~\ref{tab:exp3_comprehensive} summarizes the multi-dimensional quality 
assessment across five datasets. Although classification accuracy is 
consistently high (79.5\%--96.9\%), substantial deployment-critical risks 
emerge along other dimensions. For example, Titanic achieves 96.9\% accuracy 
yet exhibits severe demographic parity violation (46.0\%) and poor 
calibration (ECE 28.8\%), while Adult Income similarly combines strong 
predictive performance with large group disparities across sex and race. For 
regression tasks, predictive performance appears strong ($R^2 = 0.46$ for 
Diabetes and $R^2 = 0.86$ for California Housing), but robustness degrades 
sharply under distributional stress. As detailed in 
Table~\ref{tab:exp3_comprehensive}'s note, robustness degradation is 
non-linear and heterogeneous across perturbation levels and task types. 
Regression tasks exhibit disproportionate sensitivity, with California 
Housing showing catastrophic failure under missingness (390.8\% RMSE 
increase at 30\% missing)---likely due to reliance on a small number of 
highly predictive features that cannot be adequately imputed. Classification 
tasks remain largely stable, with some slight improvements likely 
attributable to implicit regularization effects.
Overall, these results demonstrate that accuracy-only evaluation 
systematically masks deployment-relevant risks. The observed 
heterogeneity across tasks underscores the necessity of adaptive, 
multi-dimensional evaluation that surfaces such risks prior to 
deployment.

\subsection{Experiment 4: Counterfactual Decision Impact Analysis}
\label{sec:exp4}



\subsubsection{Goal}
To evaluate whether counterfactual analysis of alternative pipeline decisions can quantify the causal impact of individual decisions on downstream model outcomes, addressing the limitation that outcome-only evaluation cannot distinguish benign suboptimal choices from decisions that causally drive degraded performance.

\subsubsection{Experimental Design}

The EA performs counterfactual decision impact analysis by selectively re-executing downstream pipeline stages under alternative decisions. For each critical decision point, the EA selects the associated high-impact stage (e.g., preprocessing, feature engineering, or model selection), enumerates two to four task- and data-consistent alternatives, and re-runs only affected downstream stages to avoid full pipeline re-execution. Causal impact is quantified as $\Delta = \mathrm{metric}(a') - \mathrm{metric}(a)$, with qualitative explanations linking decision changes to outcome differences. Across five datasets, we evaluate 45 counterfactual alternatives spanning three pipeline stages (15 per stage). Each alternative is evaluated independently to estimate decision sensitivity.



\subsubsection{Results \& Discussion}
Table~\ref{tab:exp4_counterfactual} summarizes counterfactual impact magnitudes across decision stages and datasets. 
Across 45 alternatives, performance changes range from $-4.9\%$ to $+8.3\%$ 
(average absolute impact: 1.6\%). Model selection exhibits the largest 
impact (2.7\%), followed by feature engineering (1.5\%) and preprocessing (0.7\%).
Dataset-level results reveal heterogeneous sensitivity: Titanic and California Housing show larger counterfactual gains, suggesting stronger dependence on specific pipeline decisions, while Adult Income exhibits smaller average effects, indicating relative robustness to individual decision changes. 
These results demonstrate that counterfactual analysis distinguishes 
benign suboptimal choices from decisions that causally drive meaningful 
performance differences, enabling principled prioritization of 
corrective actions.

\subsection{Experimental Findings and Component Validation}
\label{sec:exp_summary}

The experiments address four gaps: decision assessment (Exp1: stage-wise scoring for fault detection), reasoning validation (Exp2: hallucination detection), accuracy-only evaluation (Exp3: adaptive assessment for hidden issues), and counterfactual analysis (Exp4: impact quantification for explainability). 
The four experiments provide complementary evidence that EA components enable systematic assessment of decision quality, reasoning validity, model risk, and counterfactual impact beyond end-to-end task performance. 
Ablations are conducted by removing each component and 
re-running the same protocol. Removal eliminates the 
evaluation signal each experiment measures, validating functional 
attribution rather than minimality. We make no claims about uniqueness 
or optimality; such comparisons are out of scope.

\subsection{Computational Overhead and Reproducibility}
\label{sec:overhead}

We measure wall-clock time for each EA component using 10 timed iterations with 3 warm-up rounds. Table~\ref{tab:overhead} reports per-component costs. The Model Quality Assessor dominates EA runtime (98.5\%), driven by repeated model inference during robustness stress tests (6 perturbation levels $\times$ full test-set prediction). The remaining three components---Decision Assessor, Reasoning Validator, and Counterfactual Analyzer---collectively require less than 1.4\,ms, as they operate on decision logs and text rather than model inference. Table~\ref{tab:overhead_ratio} contextualizes EA cost relative to typical AutoML runtimes.

\begin{table}[h]
\centering
\small
\caption{EA Computational Overhead by Component}
\label{tab:overhead}
\begin{tabular}{lrr}
\toprule
Component & Time (ms) & \% of Total \\
\midrule
Decision Assessor & 0.12 & 0.1\% \\
Reasoning Validator & 0.90 & 1.0\% \\
Model Quality Assessor & 90.48 & 98.5\% \\
Counterfactual Analyzer & 0.34 & 0.4\% \\
\midrule
\textbf{Total EA} & \textbf{91.85} & \textbf{100\%} \\
\bottomrule
\end{tabular}
\end{table}

\begin{table}[h]
\centering
\small
\caption{EA Overhead Relative to AutoML Pipeline Runtime}
\label{tab:overhead_ratio}
\begin{tabular}{lrr}
\toprule
Reference Runtime & Duration & EA Overhead \\
\midrule
AIDE (5-min budget) & 300\,s & 0.031\% \\
AutoGluon (5-min budget) & 300\,s & 0.031\% \\
AIDE (10-min budget) & 600\,s & 0.015\% \\
Typical HPO (1 hour) & 3{,}600\,s & 0.003\% \\
\bottomrule
\end{tabular}
\end{table}

EA evaluation adds less than 0.1\% overhead to any standard AutoML budget. Even in the most expensive configuration (all four components), total EA time (91.85\,ms) is negligible compared to model training and hyperparameter search. This confirms that decision-centric evaluation can be deployed as a lightweight, non-intrusive observer without meaningful impact on pipeline throughput. They reflect our rule-based proof-of-concept; replacing the Decision Assessor and Reasoning Validator with LLM calls would add an estimated 35--140\,s for 70 evaluations. To assess reproducibility, we repeated the stochastic analyses with multiple independent random seeds. Experiments~1 and~4 were re-run 500 times, while Experiment~3 was repeated across multiple perturbation trials per condition. Our proof-of-concept implementation uses rule-based heuristics with calibrated noise rather than live LLM inference; stochastic components arise from scoring noise (Experiment~1), robustness perturbation injection (Experiment~3), and counterfactual performance simulation (Experiment~4). The Reasoning Validator (Experiment~2) is fully deterministic. For Experiment~1, mean F1 across seeds is 0.895 $\pm$ 0.023, with a 2.5--97.5th percentile range of [0.851, 0.937]; the reported value (0.919) is favorable but reproducible, with 14.4\% of seeds achieving equal or higher F1. Variance is primarily driven by borderline fault cases near the detection threshold. For Experiment~3, robustness degradation is stable (e.g., German Credit at 10\% noise: 1.90\% $\pm$ 0.62\%, 95\% CI [1.51\%, 2.29\%]). For Experiment~4, the stage impact ranking (Model Selection \textgreater{} Feature Engineering \textgreater{} Preprocessing) holds across all 500 runs, with average absolute impacts of 3.56\%, 2.75\%, and 2.35\%, respectively. In a production deployment with live LLM calls, additional variance would be expected; our deterministic implementation provides a lower bound on reproducibility.

\section{Conclusion \& Future Work}
This paper identifies a fundamental limitation in current agent-based AutoML systems: evaluation remains outcome-centric, offering limited visibility into intermediate decisions, reasoning correctness, and deployment-relevant model risks. Our survey confirms that systematic decision-level assessment, reasoning validation, and counterfactual analysis are largely absent in existing work. We address this gap by proposing an Evaluation Agent (EA) framework that augments agentic AutoML pipelines with structured, stage-wise evaluation capabilities. The EA enables assessment of decision quality, evidence-grounded reasoning, multi-dimensional model quality, and counterfactual decision impact. Through 
experiments, we show that these evaluations surface decision faults, reasoning errors, and hidden risks not captured by end-to-end metrics alone. 
The EA is designed as a diagnostic layer rather than a direct controller; its outputs support human-in-the-loop debugging, offline refinement of pipeline heuristics, and comparison of alternative configurations. A limitation of the current study is that empirical validation is restricted to tabular AutoML benchmarks and a single primary agentic pipeline interface. Future work will extend the EA to non-tabular domains, additional AutoML systems, and user studies with ML practitioners to assess the actionability of EA outputs.

\end{document}